\documentclass[final]{cvpr}

\usepackage{times}
\usepackage{epsfig}
\usepackage{graphicx}
\usepackage{amsmath}
\usepackage{amssymb}
\usepackage{mathtools}
\usepackage{multirow}
\usepackage{booktabs}
\usepackage{amsfonts}
\usepackage{nicefrac}
\usepackage{microtype}
\usepackage{mathrsfs}
\usepackage{soul}
\usepackage{comment}
\usepackage{rotating}
\usepackage{mwe}
\usepackage{color}
\usepackage{url}
\usepackage{diagbox}
\usepackage[pagebackref=true,breaklinks=true,colorlinks,bookmarks=false]{hyperref}

\begin{document}

\title{ClaRe: Practical \underline{Cla}ss Incremental Learning By \\ \underline{Re}membering Previous Class Representations}

\author{Bahram Mohammadi \\
Sharif University of Technology \\
{\tt\small bmohammadi@alum.sharif.edu}
\and
Mohammad Sabokrou \\
Institute For Research In Fundamental Sciences (IPM) \\
{\tt\small sabokro@ipm.ir}
}

\maketitle

\begin{abstract}
   This paper presents a practical and simple yet efficient method to effectively deal with the catastrophic forgetting for Class Incremental Learning (CIL) tasks. CIL tends to learn new concepts perfectly, but not at the expense of performance and accuracy for old data. Learning new knowledge in the absence of data instances from previous classes or even imbalance samples of both old and new classes makes CIL an ongoing challenging problem. These issues can be tackled by storing exemplars belonging to the previous tasks or by utilizing the rehearsal strategy. Inspired by the rehearsal strategy with the approach of using generative models, we propose ClaRe, an efficient solution for CIL by remembering the representations of learned classes in each increment. Taking this approach leads to generating instances with the same distribution of the learned classes. Hence, our model is somehow retrained from the scratch using a new training set including both new and the generated samples. Subsequently, the imbalance data problem is also solved. ClaRe has a better generalization than prior methods thanks to producing diverse instances from the distribution of previously learned classes. We comprehensively evaluate ClaRe on the MNIST benchmark. Results show a very low degradation on accuracy against facing new knowledge over time. Furthermore, contrary to the most proposed solutions, the memory limitation is not problematic any longer which is considered as a consequential issue in this research area.
\end{abstract}

\vspace{-3mm}
\section{Introduction}
Real-world machine learning approaches not only must be able to perfectly learn the desired concept (\textit{e.g.}, training data samples), but also should be adaptable to learn new data classes continually and gradually. For instance, robots usually face new knowledge (\textit{i.e.}, tasks) which are essential to be learned constantly \cite{mi_generalized_2020}. In such settings, Class Incremental Learning (CIL) is indispensable to properly deal with data from unknown classes. It is of considerable importance especially when the Artificial Intelligence (AI) agent is confronted with the lack of computational or memory capacity \cite{belouadah_scail_2020}. Therefore, the unavailability of the previous data instances is probable and needs to be addressed in order to have an effective method in this field of research. CIL has drawn the attention of many researchers in recent years due to its crucial role in real-world AI applications. 

\begin{figure*}[t]
	\center
	\includegraphics[width=1\linewidth,scale=0.1]{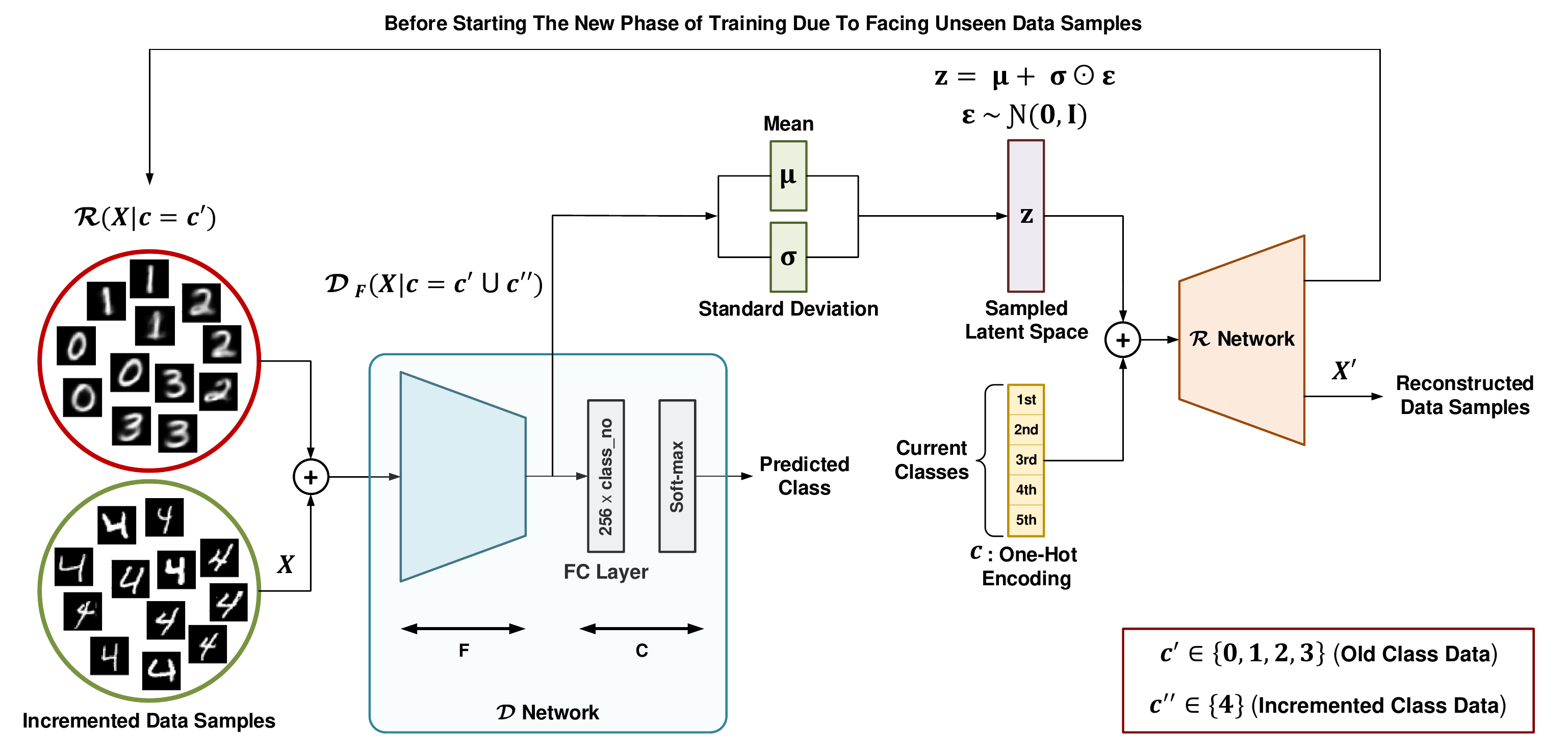}
	\caption{The outline of our proposed method, ClaRe, in the training phase which includes both $\mathcal{D}$ and $\mathcal{R}$ networks. $\mathcal{D}$ consists of a feature extractor ($\mathcal{D}_{F}$) alongside a FC network ending with a soft-max layer on top forming $\mathcal{D}_{C}$. $\mathcal{D}$ has a duty to determine which class the input data belongs to. Moreover, $\mathcal{R}$ intends to generate the old data via remembering the representations have been stored so far. Three goals are pursued during the training process: (1) minimizing the reconstruction error, \textit{i.e.}, $X \approx X'$, (2) enforcing the distributions returned by $\mathcal{D}_{\text{F}}$ to be close to a standard normal distribution ($\mathcal{N}(\mu_{z},\sigma_{z})$) and (3) enabling $\mathcal{D}$ to precisely distinguish between data instances of different classes. All of the distributions are conditioned with $c$ variable to uniquely specify the class of input samples. This approach improves the performance of $\mathcal{D}$ while $\mathcal{R}$ is capable of simulationary generating more real data to overcome the catastrophic forgetting as the most challenging problem of CIL.
	}
	\label{fig:overview}
\end{figure*}

Previously proposed state-of-the-arts can be fallen into three major categories \cite{zhao_maintaining_2020}: (1) parameter control, (2) knowledge distillation and (3) rehearsal. The key idea of parameter control strategy is to confine the learned parameters (\textit{i.e.}, weights) of the old model in confronting with learning new tasks \cite{huang_densely_2017}. Since the importance of parameters is difficult to measure accurately in a series of tasks, this strategy usually performs poorly in CIL. Knowledge distillation is another solution which is widely-used to cope with the catastrophic forgetting in the CIL problem. In this category, the main knowledge is transferred to a student model from a teacher model by facing new data. Whereas, the capabilities of the old model are appropriately preserved \cite{li_learning_2018}. The rehearsal strategy attempts to alleviate the catastrophic forgetting by utilizing the original data to reproduce the training set. Most of the methods belonging to this strategy store a limited number of original training samples to retrain the deep neural network which should learn new classes \cite{hou_learning_2019}. Generating original instances based on a generative process \cite{wu_memory_2018, zhu_unpaired_2017}, \textit{e.g.}, Generative Adversarial Networks (GANs) \cite{goodfellow_generative_2014} that become a hot research topic in AI \cite{kang_effective_2020}, is another approach of the rehearsal strategy. However, generative models prefer producing previous samples to storing them and thus this approach highly relies on the quality of the generated data. Moreover, training an additional generative model simultaneously is considered expensive for an algorithm. Note that some other solutions by the combination of the above-mentioned strategies are presented by researchers. For example, \cite{rebuffi_icarl_2017} has applied both the distillation and the rehearsal strategies. Also, \cite{castro_end_2018} has exploited these two strategies along with using a balanced fine-tuning to alleviate class imbalance.

To overcome the aforementioned challenges such as catastrophic forgetting, imbalance data and memory limitation, we propose a novel method, ClaRe, to effectively tackle the issue of catastrophic forgetting when acquiring and accumulating new knowledge is necessary. The key idea of ClaRe is to remember a lightweight representation for each class that must be learned over time while storing none of the original data in each increment is required at all. Accordingly, our proposed method is not restricted by the memory limitation. Basically, ClaRe is one neural network constituting of two relevant sub-modules, one of which is responsible for detecting the correct class of each input sample and the other memorizes precise representations for each of classes the model is confronted with. These networks support each other to play their roles in the best way. The proposed solution is very close and related to the rehearsal strategy with the approach of using generative models which is considered very promising for CIL tasks. Although ClaRe is also implementable by the conditional GAN, we merely exploit one neural network formed by two sub-modules that are trained simultaneously. Whereas using conditional GAN necessitates training two different neural networks in order to concurrently learn the concept of the input data in addition to a classifier for the task of classification.

In a nutshell, the most important contributions of ClaRe are: (1) solving the catastrophic forgetting problem caused by unavailability of the original training data, (2) providing generalization due to producing different data instances of one class, (3) imbalance class issue does not exist any longer owing to the optional number of simulated data that can be generated by our model, (4) overcoming the probable memory limitation in so many real-world applications and (5) results confirm the effectiveness of ClaRe, since it behaves very stable after adding new classes (data samples).

\section{Proposed Method}
ClaRe is a practical and effective solution for incrementally and continually learning new knowledge in such a way that there is no need to store the old data belonging to classes have been learned thus far. To this end, the main concept of original classes should be memorized. Generally, the proposed method is composed of two sub-networks: (1) $\mathcal{D}$iscriminator and (2) $\mathcal{R}$emembering network. $\mathcal{D}$ tends to correctly differentiate the input data samples and it also helps $\mathcal{R}$ to accurately memorize the representations. At the same time, The other network ($\mathcal{R}$) attempts to remember the concept of the leaned classes prior to the beginning of each training phase (except the first stage) and to memorize the representation of each class during the training process. Inspired by the Conditional Variational Auto-Encoder (CVAE) \cite{sohn_learning_2015}, these two neural networks are jointly learned on available training samples. When the trained model faces new classes, data instances following the distribution of the previous class data are generated by $\mathcal{R}$. $\mathcal{D}$ is subsequently retrained on the generated samples along with the data of new classes. A sketch of ClaRe is depicted in Fig. \ref{fig:overview}. $\mathcal{D}$ and $\mathcal{R}$ networks are explained in the following subsections.

\subsection{$\mathcal{D}$ Network}
Distinguishing between samples of various classes is the main duty of $\mathcal{D}$. This network comprises two major parts, a Feature extractor (F) and a Classifier (C). From now on, F and C are denoted by $\mathcal{D}_{F}$ and $\mathcal{D}_{C}$, respectively. $\mathcal{D}_{F}$ is constituted of two layers with the sizes of $512 \times 1$ and $256 \times 1$. $\mathcal{D}_{F}$ is trained in interaction with $\mathcal{R}$ to improve the quality of the representations should be stored by our model for each class. $\mathcal{D}_{C}$ consists of a Fully-Connected (FC) layer with the size of $256 \times class\_no$ which ends with a soft-max layer on top. $class\_no$ is the total number of classes including both old and new ones.

\begin{table*}
	\centering
	\renewcommand{\arraystretch}{1.1}
	\caption{Classification accuracy of our method, ClaRe, in the groups of $g = 1,~2~or~5$ classes at a time.}
	\label{tab:classification_accuracy}
	\begin{tabular}{|c|c|c|c|c|c|c|c|c|c|c|}
		\hline
		
		\multirow{2}{*}{$g$} & \multicolumn{10}{c|}{$Accuracy~(\%)$} \\
		
		\cline{2-11}
		
		\multirow{2}{*}{} & Class 0 & Class 1 & Class 2 & Class 3 & Class 4 & Class 5 & Class 6 & Class 7 & Class 8 & Class 9 \\
		
		\noalign{\hrule height 1.5pt}
		
		\hline
		
		1 & 100 & 99.9 & 98.6 & 95.2 & 93.4 & 89.5 & 87.6 & 83.5 & 81.3 & 78.6 \\
		
		\hline
		
		2 & \multicolumn{2}{c|}{99.9} & \multicolumn{2}{c|}{96.7} & \multicolumn{2}{c|}{95.6} & \multicolumn{2}{c|}{91.8} & \multicolumn{2}{c|}{88.2} \\
		
		\hline
		
		5 & \multicolumn{5}{c|}{96.3} & \multicolumn{5}{c|}{92.7} \\

		\hline
	\end{tabular}
	\vspace{-3mm}
\end{table*}

\subsection{$\mathcal{R}$ Network}
$\mathcal{R}$ is a two-layer neural network with the sizes of $256 \times1 $ and $512 \times 1$. The input layer of $\mathcal{R}$ is imposed to follow a specific type of normal distribution. Note that the last FC layer of $\mathcal{D}_{F}$  is equivalent to the input layer of $\mathcal{R}$. In this way of approaching the problem, $\mathcal{R}$ is capable of generating the original samples that have already been classified by $\mathcal{D}$ (see the subsection \ref{sec:training} for more details about the training process). These two neural networks as a whole, \textit{i.e.}, $\mathcal{R} + \mathcal{D}_{F}$, can be considered as an encoder-decoder neural network which is trained to generate previously seen but unavailable instances by the learned model. Nevertheless, applying a minor alternation to the encoding-decoding process is necessary to introduce some regularization of the latent representation. Consequently, an input is encoded as a distribution over the latent space instead of a single point. In this case, data generation can be done by taking a point randomly from that latent space and decode it to get new content. It is worth mentioning that, generalizability which is one of the most important features for simulationary generating previous class data, is also met by taking this solution, since for each class generated samples are not the same.

\subsection{Training ClaRe: $\mathcal{D} + \mathcal{R}$}
\label{sec:training}
$\mathcal{D}_{F}$ and $\mathcal{R}$ are trained jointly and simultaneously. An overview of ClaRe in the training stage is illustrated in Fig. \ref{fig:overview}. As stated previously, inspired by CVAEs, $\mathcal{D}_{F}$ and $\mathcal{R}$ acts as an encoder-decoder network. As a consequence, $\mathcal{D}_{F}$ is able to precisely learn the concept of different input classes and $\mathcal{R}$ memorizes their representations during the training procedure. In this case, $\mathcal{R}$ is able to remember the old unavailable data and then generates them just before the start of the training stage to facilitate this process as well as enhance the detection accuracy. Note that the training process of $\mathcal{D}_{F} + \mathcal{R}$ at the first step slightly differs from the rest. From the second step on, we need to simulate the prior training data that is not the case in the first phase. Therefore, they are able to generate appropriate amount of data. This helps us to cope with the imbalance data problem. The training of $\mathcal{D} + \mathcal{R}$ is done by optimizing the
\begin{align}
\label{eq:total_loss}
\mathcal{L}_{Total} = \mathcal{L}_{\mathcal{D}} + \mathcal{L}_{\mathcal{D}_{F}+\mathcal{R}}
\end{align}
The loss function that is minimized during the training of $\mathcal{D}_{F} + \mathcal{R}$ (\textit{i.e.}, $\mathcal{L}_{\mathcal{D}_{F} + \mathcal{R}}$) is constituted of a reconstruction term on the final layer of $\mathcal{R}$ that makes the encoding-decoding scheme as efficient as possible and a regularization term on the latent space, that has a tendency to regularize the organization of the latent representation by making distributions returned by $\mathcal{D}_{F}$, close to a standard normal distribution. This regularization term is expressed as the Kullback-Leibler (KL) divergence between the returned distribution and a standard Gaussian. In fact, we need to have control on data generation process to create data form a specific class. To this end, the label of the input data must be taken into account. Hence, $\mathcal{D}_{F}$ models the latent space ($z$) based on the input data ($X$) and another variable ($c$), \textit{i.e.}, all of the distributions are conditioned with a variable $c$. Accordingly, for each possible value of $c$, there is a $\mathcal{R}(z|X,c)$. $\mathcal{L}_{\mathcal{D}_{F}+\mathcal{R}}$ is calculated by
\begin{equation}
\small
\begin{split}
\label{eq:vae_loss}
\log & \mathcal{R}(X|c) - KL(\mathcal{D}_{F}(z|X,c)\|\mathcal{R}(z|X,c)) = \\
& \underbrace{\mathbb{E}_{z \sim \mathcal{D}_{F}} [\log{\mathcal{R}(X|z,c)}]}_{\text{Reconstruction Error}} + \underbrace{KL(\mathcal{D}_{F}(z|X,c)\|\mathcal{R}(z|c))}_{\text{Regularization}}
\end{split}
\end{equation}
Where $z \sim \mathcal{N}(\mu_{z},\sigma_{z}) = \mathcal{D}_{F}(X)$ and $c$ is a one-hot encoding to determine each class individually. When the training procedure is completed successfully, $\mathcal{D}$ is able to accurately classify the input instances and $\mathcal{R}$ can effectively simulate the old data from the second phase of training onward.

\subsection{Testing ClaRe: $\mathcal{D}$}
For the evaluation of our work, merely $\mathcal{D}$ including both major parts, \textit{i.e.}, $\mathcal{D}_{F}$ and $\mathcal{D}_{C}$, is exploited to classify the input data which is present in the test set. Note that remaining modules are not useful until the next training stage when the model is confronted with samples from new classes. Thus, our model should be retrained on both new data and generated samples.

\section{Experiments}
In this section, a thorough evaluation is carried out to show the high performance of ClaRe. The scenario and the setting for assessing our approach are provided in the subsection \ref{sec:res}. ClaRe has been implemented using Pytorch\footnote{https://pytorch.org} on GOOGLE COLAB\footnote{https://colab.research.google.com}.
The results confirm that the proposed method outperforms the other state-of-the-arts. 

\subsection{Experimental Setup}
\label{sec:res}
ClaRe is evaluated on the MNIST benchmark \cite{mnist}. Generally, learning classes incrementally by a reasonable step is very common among the proposed methods for the evaluation process. Therefore, our model learns all 10 classes in groups of $g = 1,~2~or~5$ classes at a time. For example, when $g = 1$ in the test scenarios, our model faces new classes one by one, \textit{i.e.}, one class is added in each increment. The evaluation metric is the multi-class classification accuracy on the test set while the best results are reported in each experiment. 

\vspace{-3mm}
\subsection{Experimental Results}
Table \ref{tab:classification_accuracy} shows the final results of our method. As can be seen, ClaRe is very stable against facing new knowledge or tasks owing to the very little degradation on accuracy after learning new class data. The reported results show the high functionality of $\mathcal{R}$ to generate old samples belonging to the prior classes. The insensitivity of our work to the increase of classes makes ClaRe more scalable and also more reusable.  

To show the superiority of our proposed method, we compare ClaRe with the baselines and also the other stat-of-the-arts which has been presented by \cite{wu_memory_2018}. As upper and lower bounds, joint training with all data (\textit{i.e.}, non-sequential) and sequential fine tuning are considered. Additionally, there are two other state-of-the-art methods to make a comparison with our work comprising the adaptation of elastic weight consolidation to conditional GANs proposed by \cite{seff_continual_2017} and the deep generative replay module of \cite{shin_continual_2017} implemented as an unconditional GAN followed by a classifier to predict the label. sequential fine tuning completely forgets previous tasks in the dataset, while we can observe different degrees of forgetting in the other solutions. Table \ref{tab:avg_accuracy_comparison} shows the average classification accuracy of each baseline alongside the other approaches including ClaRe. Obviously, the accuracy of our work is lower than joint training, since it shows the best possible result (upper bound) for CIL task in our test scenario while it outperforms the other state-of-the-arts by a large margin. Results confirm the high performance of ClaRe for both 5 and 10 sequential tasks. The term of sequential tasks refers to learning classes incrementally. Indeed, classes are added one by one in each increment and then the accuracy is calculated. Finally, the average value is reported for 5 and 10 classes. As it is clear in this table, our method is very close to the upper bound especially when the first five data classes are considered.

\begin{table}
    \small
	\centering
	\renewcommand{\arraystretch}{1.1}
	\caption{Average classification accuracy comparison with baselines along with the other state-of-the-art methods for 5 and 10 sequential tasks. The best result is boldface underlined.}
	\label{tab:avg_accuracy_comparison}
	\begin{tabular}{ccc}
		\hline
		
		\multirow{3}{*}{Method} & \multicolumn{2}{c}{$Accuracy~(\%)$} \\
		
		\multirow{3}{*}{} & 5 tasks & 10 tasks \\
		
		\multirow{3}{*}{} & (0-4) & (0-9) \\
		
		\noalign{\hrule height 1.5pt}
		
		Joint training (upper bound) & 97.7 & 96.9 \\	
		
		Sequential fine Tuning (lower bound) & 19.9 & 10.1 \\
		
		\hline
		
		Elastic weight consolidation \cite{seff_continual_2017} & 70.6 & 77.0 \\
		
		Deep generative replay \cite{shin_continual_2017} & 90.4 & 85.4 \\
		
		ClaRe (ours) & \textbf{\underline{97.4}} & \textbf{\underline{90.8}} \\
		
		\hline
	\end{tabular}
	\vspace{-4mm}
\end{table}

\section{Complexity}
As previously discussed, our proposed method is also implementable using conditional GAN. Therefore, we need to concurrently train two individual networks, generator and discriminator, in competing with each other. Moreover, training a classifier is essential to classify the input data. Instead of training three different modules, ClaRe is merely constituted of one neural network which includes two sub-modules, $\mathcal{D}$ and $\mathcal{R}$. Although ClaRe has an outstanding performance in both providing the model with high quality generated old samples and classifying the input data, the process of simulating original instances is time-consuming. Designing a module to effectively learn the feature space of the input data improves the model in terms of accuracy as well as the time complexity. In this case, confronting with new knowledge necessitates modifying the feature space so that it embraces the concept of new classes. However, introducing such modules can be considered very challenging and needs to be thoroughly investigated as a promising future work.

\section{Conclusion}
In this paper, we propose a practical and novel method, ClaRe, to effectively address the catastrophic forgetting problem when our model faces new knowledge or unseen concepts. ClaRe is composed of two correlated sub-modules, $\mathcal{D}$iscriminator and $\mathcal{R}$emembering network. Furthermore, $\mathcal{D}$ consists of two major parts, a feature extractor and a classifier which are known as $\mathcal{D}_{F}$ and $\mathcal{D}_{C}$, respectively. Note that $\mathcal{D}_{F}$ and $\mathcal{R}$ are trained jointly. $\mathcal{D}$ is responsible for distinguishing between the existing classes while $\mathcal{R}$ remembers the representations of previous data classes to appropriately generate old samples. The reported results confirm the superiority of our method compared to the other state-of-the-arts.

\bibliographystyle{ieee_fullname}
\bibliography{references}

\end{document}